\documentclass{INTERSPEECH2023}
\usepackage[english]{babel}
\usepackage{multirow}
\usepackage{mathtools}
\usepackage{float}
\usepackage{algorithm2e}
\usepackage[fontsize=9pt]{fontsize}
\usepackage{bold-extra}
\usepackage[T1]{fontenc}
\usepackage[subtle]{savetrees}
\usepackage[tableposition=top]{caption}

\usepackage{amssymb}
\usepackage{pifont}
\usepackage{url}
\usepackage{hyperref}
\newtheorem{definition}{Definition}

\SetKw{Continue}{continue}
\SetKw{Break}{break}

\RestyleAlgo{ruled}

\interspeechcameraready

\title{Blank Collapse: Compressing CTC emission for the faster decoding}
\name{Minkyu Jung$^1$, Ohhyeok Kwon$^2$, Seunghyun Seo$^2$, Sunshin Seo$^2$}
\address{
  $^1$Channel Corporation, Republic of Korea\\
  $^2$Naver Cloud, Republic of Korea}
\email{lep@snu.ac.kr, \{k.oh, real.seunghyun.seo, sunshin.seo\}@navercorp.com}

\begin{document}

\maketitle
 
\begin{abstract}
Connectionist Temporal Classification (CTC) model is a very efficient method for modeling sequences, especially for speech data. In order to use the CTC model as an Automatic Speech Recognition (ASR) task, the beam search decoding with an external language model like n-gram LM is necessary to obtain reasonable results. 

In this paper, we analyze the blank label in CTC beam search deeply and propose a very simple method to reduce the amount of calculation resulting in faster beam search decoding speed. With this method, we can get up to 78\% faster decoding speed than ordinary beam search decoding with a very small loss of accuracy in LibriSpeech datasets. 

We prove this method is effective not only practically by experiments but also theoretically by mathematical reasoning. We also observe that this reduction is more obvious if the accuracy of the model is higher.
\footnote{Code and examples are available at \\ \url{https://github.com/minkjung/blankcollapse}}

\end{abstract}

\section{Introduction}
\label{sec:intro}
Recently, there has been a remarkable improvement in Automatic Speech Recognition (ASR) system thanks to the End-to-End (E2E) training approaches. Especially, most ASR models have one of three popular models as a base architecture, Connectionist Temporal Classification (CTC) based model~\cite{graves2006connectionist}, recurrent neural network (RNN) transducer-based model~\cite{graves12, zhang2020transformer} and Attention-based Encoder-Decoder (AED) model~\cite{chan2016listen, dong2018speech}.

Unlike other models, CTC encodes the waveform with sequential models like Recurrent Neural Network (RNN)\cite{graves2014towards} or Transformer\cite{baevski2020wav2vec,gulati2020conformer} and outputs the probability or logit for each sample frame. Since it does not have any decoder module itself and assumes conditional independence with respect to time, it can make the recognition result directly from its output. Moreover, this model tries to learn the frame-wise phoneme representation, it originally lacks temporal information. Therefore CTC-based model is often used with beam search decoder and external Language Model (LM) like n-gram LM\cite{synnaeve2020end,baevski2020wav2vec} or deep learning-based LM~\cite{hannun2014first}. However, even though the beam search decoder shows a better result, it requires a longer decoding time. Both accuracy and decoding time is too important to give up so usually we have to take the saturation point between this trade-off. 

There have been lots of efforts to reduce the beam search decoding time so far. In \cite{freitag2017beam}, they introduced beam threshold pruning which prunes some beam candidates with relatively low scores and showed significant improvements in beam search speed. \cite{seki2019vectorized} invented vectorized CTC which also shows big enhancements in CTC-Attention-based beam search decoding while \cite{jain2019rnn} proposed an improved RNN-T beam search achieving decoding time reduction. Especially for CTC, there is a special label \textit{blank} that has multiple functions in the decoding process and the calculation of CTC loss. \cite{chen2016phone, zhang2021tiny} proposed phone synchronous decoding (PSD) in WFST which can skip the unnecessary decoding steps whose blank probability is high enough. \cite{wang2023accelerating} also tried to reduce frames with high blank probability using CTC posterior in RNN-T model training. However, simply skipping blank frames should be more careful when we consider the end-to-end character-based or subword-based CTC models since the blank symbol is essential for some cases. Some heuristics might avoid relevant problems in PSD but they are not formulated well so far.

In this paper, we research the role of blank labels deeply in the CTC beam search decoding and find an efficient way to remove the redundant computations for the blank label. In conclusion, we propose a blank collapse method for reducing the calculations in decoding which results in the improvement of decoding speed with negligible loss of accuracy. This also can be regarded as a generalization from PSD for any decoding methods and the token units, accompanied by mathematical consideration and empirical results. This can be done without any further training and is available on every CTC emission.

\begin{figure}[t!]
\centering
\includegraphics[scale=0.2]{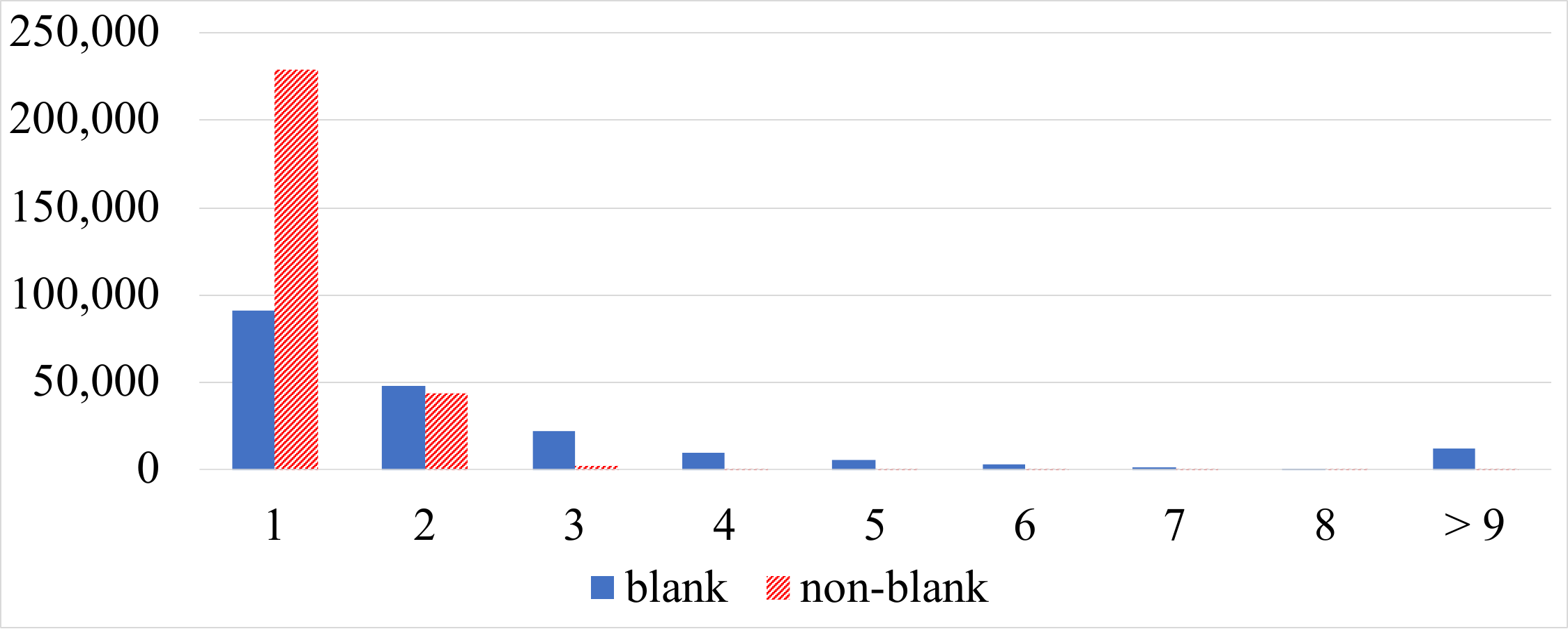}
\caption{Number of all consecutive frames for each size based on the frame type in LibriSpeech test-other. The blank type means the frame having the highest probability at the blank index while the non-blank type is for the other cases. Most of the non-blank frames last only one or two frames whereas blank frames seem to last longer.}
\label{fig:number_of_frames}
\end{figure}

\section{Analysis on CTC blank}
\label{sec:analysis}

In this section, we study the characteristics of CTC blank. Most of the notations in this section follow those in \cite{graves2006connectionist}.

\vspace{-1.5mm}
\subsection{Blanks on greedy decoding}
Before introducing the method, we have to define the blank frame. Let $\mathcal{N}_{w}(\mathbf{x}) = \mathbf{y} = (\mathbf{y}_1, \mathbf{y}_2, \cdots, \mathbf{y}_T) \in \mathbb{R}^{T \times |L'|}$ be the CTC emission probability of a certain waveform $\mathbf{x}$ from CTC model $\mathcal{N}_{w}$ where $\mathbf{y_t}=(y_t^1, y_t^2, \cdots, y_t^{|L'|}) \in \mathbb{R}^{|L'|}$ is the probability at a specified timestep $t$ and $L=\{w_1, w_2, \cdots, w_{|L|}\}$ is a set of of labels, $L'=L \cup \{\epsilon \}$, $\epsilon$ represents the blank label. CTC probability normally can be obtained by applying the softmax function to the logit from the output of LSTM or Transformer trained with CTC loss.
Each $w_l$ can be a subword or character and we use a character dictionary. 

If we want CTC greedy decoding (best path decoding), we calculate the CTC output sequence as $(\arg\max_{k \in L'} y_t^k)_{t=1}^T$ for each timestep $t$ and map the sequence with the mapping $\mathcal{B}$ in \cite{graves2006connectionist}. $\mathcal{B}$ maps a sequence of CTC outputs to a label sequence by removing all blanks and repeated labels from the sequence. Thus by the definition of $\mathcal{B}$, consecutive blanks play the same role as a single blank. (e.g. $\mathcal{B}(a \epsilon \epsilon \epsilon a \epsilon b) = \mathcal{B}(a \epsilon a \epsilon b) = aab$). This property motivates us to think that consecutive blank outputs could be replaced by a single blank output. 
Also, we note that we can't ignore a series of blanks entirely because the existence of a blank label plays an important role that used for representing a repetition of the non-blank label. That's why we have to leave at least one blank label on behalf of the following blanks.

However, consecutive blanks at the beginning and the ending of a sequence can be omitted entirely since they affect nothing to the result of greedy decoding. In other words, in greedy decoding, we are allowed to collapse the consecutive blanks to a single blank and drop all blanks before the first non-blank output and after the last non-blank output before decoding. More formally, we get the following.

\begin{definition}
\label{def:greedy}
$\mathcal{G}: \mathbb{R}^{T \times |L'|} \to \{1, ..., |L'|\}^{\le T}$ is the \textbf{CTC greedy decoding} or \textbf{best path decoding} given by $$\mathcal{G}(\mathbf{y}) = \mathcal{B}((\arg\max_{k \in L'} y_t^k)_{t=1}^T)$$ for a CTC emission probability $\mathbf{y}$. $\hat{E}_\mathbf{y}$ is called a set of \textbf{weak blank frames} of $\mathbf{y}$ defined as $$\hat{E}_\mathbf{y} \coloneqq \{t \le T : \arg \max_{k \in L'} y_t^k = \epsilon \}.$$ $\phi : \mathcal{P}(\{1, \cdots, T \}) \to \mathcal{P}(\{1, \cdots, T \})$ is a function called the \textbf{consecutive extension} defined by
$$\phi(E) \coloneqq \{t \in E : (t-1 \in E) \lor (t=1) \lor (s \in E, \forall s \ge t) \}$$
\end{definition}

By definition, it is straightforward that 
\begin{equation}\label{eq:week_collapse}
    \mathcal{G}(\mathbf{y}) = \mathcal{G}((\mathbf{y}_t)_{t \notin \phi(\hat{E}_X)})
\end{equation}
for a CTC emission probability $\mathbf{y} \in \mathbb{R}^{T \times |L'|}$. This means we may drop frames included in $\phi(\hat{E_\mathbf{y}})$ for CTC greedy decoding, even though it is not useful in practice. 




\vspace{-1.5mm}
\subsection{Blanks on CTC beam search}

\autoref{eq:week_collapse} might not be valid for the CTC beam search decoding. Let $C_t$ be a set of beam candidates for timestep $t$. During CTC beam search, we calculate $P_b(c, t)$ and $P_{nb}(c, t)$, blank probability and non-blank probability respectively for each candidate $c\in C_t$ and each timestep $t \le T$. As $t$ proceeds, $P_b$ and $P_{nb}$ have to be updated for the case of stay ($c \in C_{t+1}$) or extend to another path ($c'=c \cup \{k\} \in C_{t+1}$) according to the following rules: 
\begin{equation} \label{eq:ctc_beam_search}
\begin{aligned}
    P_b(c, t+1) &= P_{tot}(c, t) \cdot y_t^{\epsilon} \\
    P_{nb}(c, t+1) &\mathrel{+}= P_{nb}(c, t) \cdot y_t^k, \qquad \text{for } k= c_{-1}, \\
    P_{nb}(c', t+1) &\mathrel{+}= 
    \begin{dcases}
        P_{b}(c, t) \cdot y_t^{k}, &\text{ if } k = c_{-1}, \\
        P_{tot}(c, t) \cdot y_t^{k}, &\text{ otherwise,} 
    \end{dcases} \\
\end{aligned}
\end{equation}
where $P_{tot}(c, t) = P_b(c, t) + P_{nb}(c, t)$ and $c_{-1}$ represents the last label of $c$. We use $\mathrel{+}=$ instead of $=$ for non-blank probability because they can be added from another beam extension.
CTC beam search with a language model (LM) sorts the candidates by their score defined by 
\begin{equation}\label{eq:beam_search_score}
   \mathit{Score}(c, t) \coloneqq  P_{tot}(c, t) \cdot \lambda P_{LM}(c, t). 
\end{equation}
Here, $P_{LM}$ is a probability of LM and $\lambda$ is the LM weight (other hyper-parameters like a length penalty are not considered for this time). As we see \autoref{eq:ctc_beam_search} even though $y_t^\epsilon$ is the biggest probability, $y_t^k$ is applied to the non-blank probability which can be accumulated as $t$ goes. Thus collapsing \textit{weak blank frames} is too risky to collapse carelessly.

Consequently, we consider a stronger condition for the blank frame which affects search little enough to collapse.

\begin{definition}\label{def:bc}

$E_\mathbf{y}^{\theta}$ is called a set of \textbf{strong blank frames} or just \textbf{blank frames of $\theta$} for $\mathbf{y}$ defined by $$E_\mathbf{y}^\theta \coloneqq \{t \le T : y_t^\epsilon > \theta \}$$ where $\theta$ is called \textbf{blank threshold}.
\end{definition}

Compared to \textit{weak blank frames}, on \textit{blank frames of $\theta$}, it is more confident for CTC model to predict that these frames are for the blank label. Also, we can control the confidence by $\theta$ for alleviating the difference from the result from original CTC emissions. We propose our method with this definition in the next chapter followed by its reasoning of it.

\vspace{-1.5mm}
\subsection{Comparison between blank and non-blank}
\label{subsec:comparison}

Although other consecutive non-blank labels also might be collapsed into a single label in greedy decoding, we don't cover the case of non-blank labels because non-blank labels (a) don't occur often in a row and (b) are riskier than blank labels.

As we can see in \autoref{fig:number_of_frames}, non-blank usually occurs at most two frames consecutively whereas blank lasts longer once it occurs. Our ultimate purpose is to reduce the decoding time and collapsing consecutive non-blanks into a single non-blank will make little difference in decoding time.

\vspace{-1.5mm}
\section{Blank collapse method}
Now we propose a new method called \textit{blank collapse} which drops all collapsible frames before the CTC beam search in order to reduce the size of decoding frames. Additionally we study the theoretical reasoning behind this method, followed by its limitations and implementation.

\vspace{-1.5mm}
\subsection{Definition of the method}

\begin{definition}[blank collapse]
$\mathbf{F_{\theta}}: \mathbb{R}^{T \times |L'|} \to \mathbb{R}^{\le T \times |L'|}$ is called \textbf{blank collapse} method with $\theta$ defined by $\mathbf{F_{\theta}}(\mathbf{y}) = (\mathbf{y_t})_{t \in \mathbf{I_\theta}}$, for a CTC emission probability $\mathbf{y}$ where $\mathbf{I_\theta} \coloneqq \phi(E_\mathbf{y}^\theta) ^C $. We call $\phi(E_\mathbf{y}^\theta)$ a set of \textbf{collapsible frames} of $\mathbf{y}$ with $\theta$. The method using $\mathbf{\hat{I}}=\phi(\hat{E}_\mathbf{y})^C$ as an index set is called \textbf{weak blank collapse}.
\end{definition} 

In other words, from the original CTC emissions, this method drops the blank frames if they occur in the front, the last, or following another blank frame. If we drop these collapsible frames, CTC emissions will be compressed on its length, resulting in a shorter beam search time. 

PSD can be seen as a case with $\mathbf{I}=(E_\mathbf{y}^\theta)^C$ which collapses every blank frame without the consecutive extension. This is generally fine when the model uses phoneme as a unit but can cause a significant difference otherwise according to Chapter 2.1.

This method certainly has to maintain the accuracy of beam search as much as possible and this is accomplished by the definition of blank frames. Since $y_t^\epsilon > \theta$, it automatically implies $y_t^k < 1-\theta$, $\forall k \in L$. This means that we can assure that the change of the non-blank probability has to be limited on blank frames while the blank probability almost remains as same before for sufficiently large $\theta$. We can almost surely ignore these frames for this reason.

\vspace{-1.5mm}
\subsection{Limitations of the method}
Even though we can almost surely omit the consecutive blank frames, this method always as a nonzero opportunity that could harm the original result. No matter how $\theta$ is large, sometimes consecutive blank frames can change the score quite a lot and reverse the order. As \autoref{eq:ctc_beam_search} shows, the non-blank accumulates the score not only from its beam but also from another beam as they merge their scores. Namely, $$P_{nb}(c, t+1) \mathrel{+}= P_{tot}(c_{:-1}, t) \cdot y_t^{c_{-1}}$$ where $c_{-1}$ is the last label of the beam $c$ and $c_{:-1}\coloneqq c-c_{-1}$. In this equation, we can limit the scale of $y_t^{c_{-1}}$ but we don't know how $P_{tot}(c_{:-1}, t)$ be larger than the other beam scores. Again, every beam always has a chance to add a relatively large score from another beam having a much higher score. LM score can cause deepen this kind of side effect. Additionally, this tells us why we collapse only blanks not non-blanks, mentioned in section \ref{subsec:comparison} (b). Collapsing frames with high non-blank emission probability can't limit $P_{nb}$ which can cause serious distortion.

Nevertheless, this method is still effective making little difference from the original. This is because the non-blank probability still has an upper limit $\theta$ and proper \textit{beam threshold pruning} \cite{freitag2017beam} prevents the beams from overspreading. Trivially, the higher $\theta$, the less distortion we could expect.

\vspace{-1.5mm}
\subsection{Implementation}
To implement this method, we take advantage of a special utility function \textit{unique\_consecutive} from \texttt{PyTorch}\cite{NEURIPS2019_9015}. First, we get a vector $\mathbf{\Phi}$ where each value $\phi_t$ represents whether the frame at timestep $t$ belongs to the blank frames or not. This function returns each unique value of a vector in order and how many times a such value occurs consecutively. Using the returned value from this function with a vector $\mathbf{\Phi}$, we can get $\mathbf{u}, \mathbf{c}$ representing how many blank/non-blank frames occur in a row. By these values, we can leave only non-collapsible frames, $\mathbf{I}_\theta$.

\begin{figure}[h]
\includegraphics[scale=0.25]{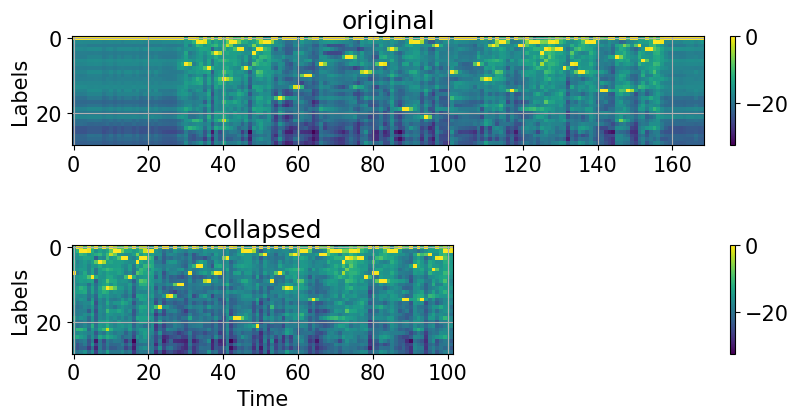}
\caption{Log probability of original/collapsed CTC emission for a sample waveform in LibriSpeech. }
\label{fig:bc}
\end{figure}

This method is also compatible with timestep alignment \cite{10.1007/978-3-030-60276-5_27} since we can reorder the timestep alignment result with $\mathbf{I_\theta}$. The detail method is described in \autoref{alg:collapse}. \autoref{fig:bc} shows the resulting image of collapsed CTC emission before and after the blank collapse. Yellow represents high log probability and the top row is for the blank label. As we can see, consecutive blank frames disappear in collapsed emission. The total length of the collapsed emission reduces from 169 to 102. The ground truth transcript is \textit{I had that curiosity beside me at this moment}.

\begin{algorithm}
\caption{blank collapse}\label{alg:collapse}
\KwData{CTC probability $\mathbf{y} \in \mathbb{R}^{T \times |L'|}$, blank threshold  $\theta$}
\KwResult{collapsed probability $\mathbf{F_\theta(y)} \in \mathbb{R}^{\le T \times |L'|}$, indices $\mathbf{I}_\theta$}
$\mathbf{\Psi} \gets (\psi_1, \cdots, \psi_T)$ where $\psi_t = [y_t^{\epsilon} \ge \theta]$, $\forall t \le T$ \;
$N_0 \gets 0$ \CommentSty{// number of first blanks} \;
$\mathbf{u}, \mathbf{c} \gets \mathit{unique\_consecutive}(\mathbf{\Phi})$ \;
$\mathbf{C} \gets [ ] $\;
$k \gets 0$ \;
\For{$i \gets 1$ \KwTo length($\mathbf{c}$)}{
  \eIf{$u_i$ is True}{
    \eIf{i = 1}{
        $N_0 \gets c_i$
    }
    {
        \eIf{i = length($\mathbf{c}$)}{
            \Break
        }
        {
            $\mathbf{C} \gets [\mathbf{C}; c_i]$    \CommentSty{// append number of blanks}
        }
    }
  }
  {
    $\mathbf{C} \gets [\mathbf{C}; [1, \cdots, 1]]$ \CommentSty{// append 1's with number of non-blanks}
  }
  $k \gets k + c_i$
}
$\mathbf{I}_\theta \gets \mathrm{cumsum}(\mathbf{C}) - 1 + N_0$ \;
$\mathbf{F_\theta(y)} \gets (\mathbf{y_t})_{t \in \mathbf{I}_\theta}$ \;
\end{algorithm}

\section{Experiments}
\label{sec:experiments}

\subsection{ASR model and datasets}
\label{subsec:baseline}

We use the {\em wav2vec 2.0} \textsc{Base} / \textsc{Large}~\cite{baevski2020wav2vec} model pre-trained on the unlabeled audio data of LibriVox dataset~\cite{librivox} and fine-tuned on either 10 minutes, 100 hours, and 960 hours of transcribed LibriSpeech dataset\cite{panayotov2015librispeech}. We do not fine-tune further and use a CTC beam search decoder and 4-gram word LM provided by \texttt{torchaudio}~\cite{yang2022torchaudio} which uses \cite{kahn2022flashlight} for the decoder. Since this decoder uses the logit of each beam for sorting beam candidates we get the probability vector explicitly by the softmax function to apply our method. Also, we try the beam threshold for $\gamma=10, 30, 50$.

Beam search decoding in an end-to-end manner uses 32 batch size, 1,500 beams, LM weight 1.57, and length penalty -0.64 for every experiment in order to simulate the experiments done by \cite{baevski2020wav2vec}. For the analysis of their effectiveness, we use various beam thresholds ($\gamma$) and blank thresholds ($\theta$). Every experiment is done on Intel(R) Xeon(R) Gold 5120 CPU @ 2.20GHz.

We evaluate our method on LibriSpeech dev and test sets. Firstly we check the proportion of collapsible frames to all frames from a CTC emission with $\theta=0.999, 0.99, 0.9$, and weak collapse (Definition \ref{def:bc}). As we can see at \autoref{table:dist}, about half of CTC frames seem to be collapsible frames. Additionally, there are more collapsible frames with the lower $\theta$ than the higher one, though the gap between them is not that big. With this observation, we can estimate that the CTC model predicts a certain frame to a blank frame with very high confidence so that there are not many frames with ambiguous certainty. 

\vspace{-1.5mm}
\begin{table}[hb!]
\caption{The percentage of collapsible frames to all frames from CTC emissions with {\em wav2vec 2.0} \textsc{Large} model fine-tuned on 960 hours.}\label{table:dist}
\centering
\begin{tabular}{|c|c|c|c|c|}
\hline
$\theta$     & 0.999 & 0.99  & 0.9   & weak  \\ \hline
dev-clean  & 42.87 & 43.24 & 43.60 & 43.97 \\ \hline
dev-other  & 42.98 & 43.77 & 44.51 & 45.20 \\ \hline
test-clean & 43.88 & 44.27 & 44.65 & 45.05 \\ \hline
test-other & 43.83 & 44.67 & 45.44 & 46.15 \\ \hline
\end{tabular}
\end{table}

\begin{table}[ht!]
\caption{Word Error Rate (WER) (\%) and Real Time Factor (RTF) with its reduction ratio compared to the original on LibriSpeech dev/test sets with $\gamma=50$. RTF includes the time spent executing blank collapse which is less than a second. }\label{table:wer}
\centering
\setlength\tabcolsep{3.5pt}
\renewcommand{\arraystretch}{1.2}
\begin{tabular}{|c|c|c|c|c|c|}
\hline
$\theta$                    &     & dev-clean & dev-other & test-clean & test-other \\ \hline
\multirow{2}{*}{original} & WER & 1.781     & 3.509     & 2.031      & 3.681      \\ \cline{2-6} 
                          & RTF & 0.278     & 0.290     & 0.279      & 0.291      \\ \hline
\multirow{2}{*}{0.999}    & WER & 1.781     & 3.511     & 2.031      & 3.681      \\ \cline{2-6} 
                          & RTF & \textbf{0.160} {\scriptsize (\textbf{0.42})}     & \textbf{0.167} {\scriptsize (\textbf{0.42})}     & \textbf{0.157} {\scriptsize (\textbf{0.44})}     & \textbf{0.165} {\scriptsize (\textbf{0.43})}     \\ \hline
\multirow{2}{*}{0.99}     & WER & 1.783     & 3.513     & 2.029      & 3.683      \\ \cline{2-6} 
                          & RTF & \textbf{0.160} {\scriptsize (\textbf{0.42})}    & \textbf{0.165}  {\scriptsize (\textbf{0.43})}   & \textbf{0.156} {\scriptsize (\textbf{0.44})}     & \textbf{0.163} {\scriptsize (\textbf{0.44})}     \\ \hline
\multirow{2}{*}{weak}     & WER & 1.866     & 3.749     & 2.109      & 3.834      \\ \cline{2-6} 
                          & RTF & 0.158 {\scriptsize (0.43)}    & 0.161 {\scriptsize (0.45)}    & 0.154 {\scriptsize (0.45)}     & 0.159 {\scriptsize (0.45)}     \\ \hline
\end{tabular}
\end{table}

\subsection{Experimental results}
\label{subsec:results}
\autoref{table:wer} shows the accuracy and decoding time of each experiment with $\gamma=50$. As we can see, blank collapse with sufficiently high $\theta$ shows a significant improvement in its decoding speed with little difference in accuracy. For test-clean, $\theta=0.99$ shows about 44\% time reduction at most, which equals to about 78\% speed enhancement without any accuracy loss. Weak blank collapse shows the best speed among all settings with not a small distortion on the accuracy. Thus we may know that sufficient $\theta$ is safe for this method. 

\begin{figure}[ht]
\centering
\includegraphics[scale=0.2]{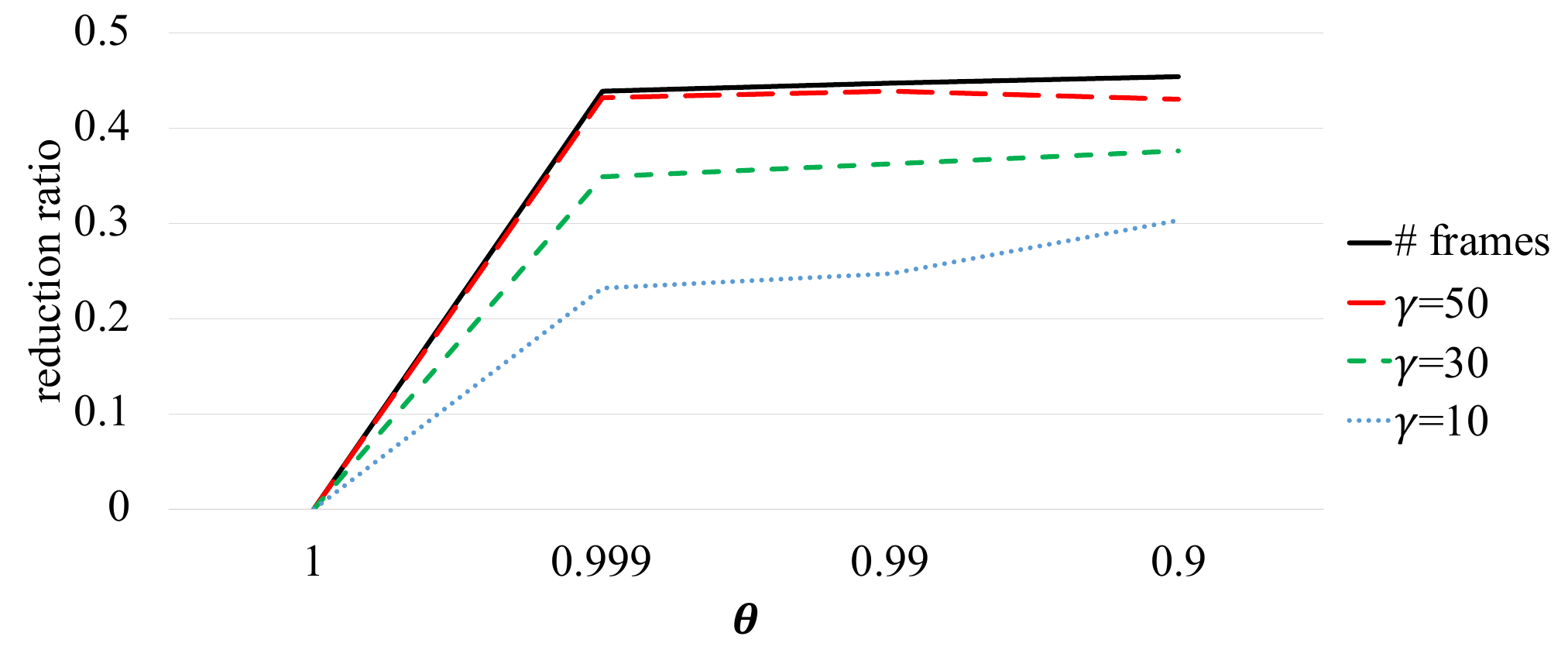}
\caption{Reduction ratio of the decoding time by blank collapse in LibriSpeech test-other according to $\theta$ for each $\gamma$, compared to the reduction ratio of the number of frames.}
\label{fig:frames_decoding}
\end{figure}

Interestingly, the reduction ratio of the decoding time seems to be almost as much as the ratio of their frame sizes, 43.2\% and 43.8\% respectively for test-other. However, this is not always true. As we see in \autoref{fig:frames_decoding}, with $\gamma=50$, the reduction ratio of decoding time is very similar to that of the size of frames. On the contrary, $\gamma=10, 30$ shows a little bit less effect than $\gamma=50$. With this phenomenon, we can estimate that the ratio of the time consuming on blank frames to that on non-blank frames is relatively lower on the smaller $\gamma$. This is because the beam threshold pruning with small $\gamma$ reduces the time more on the blank frames than the others. In other words, a higher $\gamma$ beam search takes longer time on blank frames than lower $\gamma$ because it has more candidates which must be pruned with the lower $\gamma$.

\autoref{table:by_models} shows the correlation between the model type and the improvement of decoding time by blank collapse. The larger the size of the model and the larger the dataset has been fine-tuned, the improvement of the speed gets more evident. Actually, these two factors directly affect the accuracy of the model, which is the key factor in deciding how many frames can be collapsed out. This is because the model with better accuracy tends to provide the CTC emission probability with higher confidence and it makes the blank probability high enough to be dropped. In other words, a good model can tell whether a certain frame can be collapsed out or not.

\begin{table}[ht!]
\caption{The percentage of decoding time improvement on various model types, depending on the size of the model and the fine-tuning dataset. }\label{table:by_models}
\centering
\begin{tabular}{|c|c|c|c|c|}
\hline
$\theta$                  & 0.999 & 0.99  & 0.9   & weak  \\ \hline
\textsc{Base} / 10 min  & 23.56 & 28.22 & 31.54 & 35.81 \\ \hline
\textsc{Large} / 10 min & 35.95 & 39.05 & 40.34 & 40.59 \\ \hline
\textsc{Large} / 100h   & 41.08 & 42.50 & 43.54 & 43.96 \\ \hline
\textsc{Large} / 960h   & \textbf{43.30} & \textbf{43.99} & \textbf{45.36} & \textbf{45.70} \\ \hline
\end{tabular}
\end{table}

\vspace{-1.5mm}

\vspace{-1.5mm}
\subsection{Results on other decoder settings}

In this section, we discuss proposed methods in two different decoder settings on LibriSpeech test-other subset. 
Firstly, we decode both vanilla frames and collapsed ones with a CTC beam search decoder fused with a Transformer based LM. 
We use word-level 20 decoder layers Transformer for LM\cite{synnaeve2020end,baevski2020wav2vec} and the same decoder in section \ref{subsec:baseline}.
However, unlike the n-gram fusion which shows significant improvement in inference speed with a negligible WER degradation, there is only $1\%$ gain in decoding time when $\theta$ is $0.999$.
This must be because most of the time is spent on huge LM inference when decoding with a neural LM.

Nextly, we experiment with a WFST-based beam search decoder \cite{mohri2002weighted}.
We compile the same 4-gram LM in section \ref{subsec:baseline} for a language graph and a CTC graph using k2 framework \cite{povey2021speech}. And we implement a WFST decoder using Kaldi \cite{povey2011kaldi}.
We also observe the improvement of decoding time by blank collapse with WFST-based decoding. The reduction ratio of the decoding time is about 33\% with a very small loss in accuracy similar to the case for the end-to-end decoder. Since the hyper-parameters of the WFST decoder are different from the end-to-end decoder, its improvement seems to be relatively low. If we find more optimized hyper-parameters for the WFST decoder, we expect a similar speed improvement to that of the end-to-end decoder.

\vspace{-1.5mm}
\subsection{Analysis of side-effects} 
It seems a bit strange that the reduction ratio on $\theta=0.9$ is lower than $\theta=0.999, 0.99$ at $\gamma=50$ as shown in \autoref{fig:frames_decoding}. It's rare but it also does happen on our internal dataset as well. We interpret that over-collapsing might drop some good candidates unintentionally which might lead to faster decoding with appropriate pruning for the rest of the time.

Additionally, there are some cases in which the accuracy turns out to be better when collapsed than the original. For example, WER of test-clean with $\theta=0.99$ (2.029) is lower than that of the original (2.031) though it is a small amount. We guess that blank collapse may drop some frames having harmful information potentially. However, such cases do not occur very often and the difference is usually small throughout our experiments.

\section{Conclusions}
\label{sec:conclusions}
In this paper, we analyze the characteristics of the blank label which is used as a special role in the CTC model. We define the blank frame as a frame with a high blank probability $y_t^\epsilon$ and find out that usually CTC emission has a bunch of blank frames. It is also discussed that the blank frames can be omitted in CTC beam search decoding in almost every case.

By this, we finally propose a new method called blank collapse which intends to reduce the collapsible frames in order to improve the decoding speed with minimal loss of accuracy. It is shown by several experiments that this method actually can improve the decoding speed. We also find that we can collapse more frames when the model is well-trained on a dataset and with a higher beam threshold.

As future work, we expect that our method can be plugged into any E2E ASR models using CTC loss as a regularization.






\bibliographystyle{IEEEtran}
\bibliography{mybib}

\end{document}